\documentclass{article}

\usepackage{arxiv}

\usepackage[utf8]{inputenc} 
\usepackage[T1]{fontenc}    
\usepackage{hyperref}       
\usepackage{url}            
\usepackage{booktabs}       
\usepackage{amsfonts}       
\usepackage{nicefrac}       
\usepackage{microtype}      
\usepackage{lipsum}		
\usepackage{graphicx}
\usepackage{natbib}
\usepackage{doi}

\usepackage{multirow}
\usepackage{array}
\usepackage{float}
\usepackage[dvipsnames]{xcolor}
\usepackage{amsmath}
\usepackage{booktabs} 
\usepackage[justification=centering]{caption} 
\usepackage[acronym]{glossaries} 

\usepackage{calc}
\newsavebox\PictureBox
\usepackage{authblk}

\usepackage{geometry} 
\usepackage{changepage}

\newacronym{MethodAcronym}{Rad4XCNN}{MethodAcronym}

\title{Rad4XCNN: a new agnostic method for \textit{post-hoc} global explanation of CNN-derived features by means of radiomics}

\date{April 26, 2024}	

\author[1]{Francesco Prinzi}
\author[2,$\star$]{Carmelo Militello}
\author[1]{Calogero Zarcaro}
\author[1]{Tommaso Vincenzo Bartolotta}
\author[3,2]{Salvatore Gaglio}
\author[1]{Salvatore Vitabile}

\affil[1]{Department of Biomedicine, Neuroscience and Advanced Diagnostics, University of Palermo, Palermo, 90127, Italy}
\affil[2]{Institute for High-Performance Computing and Networking (ICAR-CNR), National Research Council, Palermo, 90146, Italy}
\affil[3]{Department of Engineering, University of Palermo, Palermo, 90128, Italy}

\affil[$\star$]{Corresponding author: email: carmelo.militello@cnr.it}

\renewcommand{\headeright}{}
\renewcommand{\undertitle}{}
\renewcommand{\shorttitle}{Rad4XCNN: a \textit{post-hoc} global explanation of CNN-derived features by means of radiomics}

\hypersetup{
    pdftitle={Rad4XCNN: a new agnostic method for \textit{post-hoc} global explanation of CNN-derived features by means of radiomic},
    pdfsubject={cs.AI},
    pdfauthor={Francesco~Prinzi, Carmelo~Militello, Calogero~Zarcaro, Tommaso V.~Bartolotta, Salvatore~Gaglio, Salvatore~Vitabile},
    pdfkeywords={Explainable AI, Radiomics, Convolutional Neural Networks, Breast Cancer, Clinical Decision Support Systems},
}

\begin{document}
\maketitle

\begin{abstract}
-\textbf{Background and Objective}: In recent years, machine learning-based clinical decision support systems (CDSS) have played a key role in the analysis of several medical conditions. Despite their promising capabilities, the lack of transparency in AI models poses significant challenges, particularly in medical contexts where reliability is a mandatory aspect. However, it appears that explainability is inversely proportional to accuracy. For this reason, achieving transparency without compromising predictive accuracy remains a key challenge. \textbf{Methods}: This paper presents a novel method, namely \acrshort{MethodAcronym}, to enhance the predictive power of CNN-derived features with the inherent interpretability of radiomic features. \acrshort{MethodAcronym} diverges from conventional methods based on saliency maps, by associating intelligible meaning to CNN-derived features by means of Radiomics, offering new perspectives on explanation methods beyond visualization maps. \textbf{Results}: Using a breast cancer classification task as a case study, we evaluated \acrshort{MethodAcronym} on ultrasound imaging datasets, including an online dataset and two in-house datasets for internal and external validation. Some key results are: \textit{i)} CNN-derived features guarantee more robust accuracy when compared against ViT-derived and radiomic features; \textit{ii)} conventional visualization map methods for explanation present several pitfalls; \textit{iii)} \acrshort{MethodAcronym} does not sacrifice model accuracy for their explainability; \textit{iv)} \acrshort{MethodAcronym} provides a global explanation enabling the physician to extract global insights and findings. \textbf{Conclusions}: Our method can mitigate some concerns related to the explainability-accuracy trade-off. This study highlighted the importance of proposing new methods for model explanation without affecting their accuracy.
\end{abstract}


\section{Introduction}
\label{sec:Introduction}

The use of computer-aided tools based on Artificial Intelligence (AI) methods has increased significantly. These tools utilize machine learning and deep learning architectures in many types of Decision Support Systems (DSS). A particular class of DSS is represented by Clinical decision support systems (CDSS), implemented to support activities in critical healthcare processes. Although data-driven methods play a crucial role in CDSSs development, their use in medicine still harbors many pitfalls. New deep learning methods and the growing data availability have enabled the development of powerful but uninterpretable CDSS, and, for this reason, the field of Explainable AI (XAI) has gained considerable attention. Regulatory authorities have discussed the lack of explainability \cite{kundu2021ai, Arrieta2020}. The US Federal Trade Commission emphasizes that AI tools should be transparent, explainable and fair \cite{smith2020using}. The European Parliament has adopted the General Data Protection Regulation (GDPR), in which meaningful explanations of the logic involved are declared mandatory when automated decision-making is performed \cite{Guidotti2018}. 
Finally, with the Regulation (EU) 2024/1689 of the European Parliament, on July 2024 was approved the AI Act, establishing harmonized rules on the use of artificial intelligence-based systems \cite{EU2024ActAI}. In this perspective, the AI Act aims to achieve its stated policy objectives with a focus on transparency and human oversight \cite{panigutti2023role}.

From a practical point of view, the lack of transparency makes both doctors and patients skeptical about these new technologies. Opaque AI systems can impair the doctor-patient relationship and jeopardize patient trust \cite{amann2020explainability}. For this reason, scientific research is working on making AI transparent and/or explainable \cite{combi2022manifesto}. 
In medical contexts, several aspects can affect the model reliability and demand explainability \cite{holzinger2016interactive, holzinger2019causability}. Consequently, methods that enable transparency and explainability can help validate the models, improve the knowledge domain, and increase the use of these systems in the real world.
For these reasons, although it is not yet fully accepted \cite{Bornstein2016, Ghassemi2021, McCoy2021, london2019artificial}, explainability is increasingly becoming a requirement that these systems should fulfill \cite{Jovanovic2022}.

Deep learning architectures have demonstrated remarkable capabilities in extracting intricate patterns and features from huge datasets, enabling unprecedented predictive performance in various fields, including medical image analysis. Although deep features achieve impressive accuracy, the question of explainability is ignored and several approaches were proposed for their \textit{post-hoc} explanation. These approaches focus mainly on visualization maps, which have shown several limitations regarding their reliability. As an example, applying different saliency map methods can result in different explanations \cite{cerekci2024quantitative, prinzi2024yolo, Zhang2022overlooked}. Furthermore, while saliency maps provide a form of local explanation, it has been shown that a global explanation is necessary for clinical model validation \cite{prinzi2023explainable}.

Radiomics \cite{gillies2016radiomics, lambin2017radiomics} is a new alternative approach to introduce explainability within the feature extraction process in radiological imaging \cite{prinzi2024micro}. Radiomic uses mathematical formulas to analyze grayscale histograms, ROI shapes, or texture-defining matrices. Consequently, each radiomic feature's significance is well known, and meaningful clinical conclusions can be drawn through model explanation \cite{prinzi2024breast}. In addition, radiomics is more suitable for training in small dataset scenarios. In fact, model training relies on algorithms designed for tabular data analysis, which are more appropriate for training with limited data \cite{an2021radiomics,traverso2018repeatability}. Although radiomic features have the great properties of training in small data scenarios and inherent explainability, their predictive power seems considerably weaker than deep features, as has been shown in several works \cite{lisson2022deep, sun2020deep, truhn2019radiomic, wei2021radiomics}. 
As a result, deep learning architectures are implemented when accuracy is the metric to maximize \cite{truhn2019radiomic}.

Considering that the feature extraction process from medical images represents the main step to implement high accuracy and explainability, several studies analyzed the interpretable capability of radiomic features \cite{varriano2022explainability} and the differences between learned features and radiomic features \cite{rundo2024image}.
The dilemma concerning the trade-off between explainability and accuracy arises \cite{vanDerVeer2021trading}. Hence, the use of deep features would ensure highly performing but poorly explainable models, in contrast, radiomic features would ensure less accurate but interpretable models.

In this study, we combine the predictive power of deep features with the intrinsic explainability characteristic of radiomic features. To this aim, state-of-the-art architectures such as ResNet, DenseNet, and Vision Transformer (ViT) for deep feature extraction were employed \cite{rahman2024gliomacnn}. Simultaneously, a radiomic workflow to extract interpretable signatures was implemented. We performed a comparative analysis between these two feature sources in terms of predictive performance and explainability. 
Moreover, we propose \acrshort{MethodAcronym}, a new method leverages deep learning for enhanced performance while exploiting the inherent interpretability of radiomic features. 
In particular, from a methodological perspective, \acrshort{MethodAcronym} doesn't introduce constraints or bottlenecks that can harm accuracies in deep architecture training \cite{alvarez2018towards,elbaghdadi2020self} while it explains CNN-derived features by means of radiomics. From a clinical perspective, \acrshort{MethodAcronym} uses quantitative radiomic features for explanations, introduces a form of global explanation, enriches image interpretation with intelligible features, and contributes to improved reliability, consistency, and confidence in radiologic practices.
In addition, it provides new perspectives on the development of explanation methods not solely relying on visualization maps \cite{papanastasiou2023attention}. 
A breast cancer classification task was proposed as a case study to evaluate the effectiveness of \acrshort{MethodAcronym}. To this end, we acquired two in-house datasets, one for training and internal testing and the other for external validation.

The main contributions are:
\begin{itemize}
    \item \acrshort{MethodAcronym} a new method able to explain deep features by means of radiomics;
    \item a comparison between deep and radiomic features in terms of explainability and accuracy, providing insights on the importance of proposing new methods to overcome the accuracy-explainability trade-off dilemma;
    \item the use of deep architectures for breast cancer classification in ultrasound imaging exploiting an online dataset, and two in-house datasets for internal test and external validation;
\end{itemize}

The rest of the article is structured as follows: Section~\ref{sec:Background} 'Background' Section introduces the main concepts of explainable AI and focuses on methods for explaining deep architectures and radiomic models. Section~\ref{sec:MaterialsMethods} 'Materials and Methods' describes the datasets and the methods implemented, including our novel approach for global explanation of deep features. Section~\ref{sec:ExperimentalResults} 'Experimental Results' compares the results of deep architectures and radiomic models in terms of accuracy and explainability. Section~\ref{sec:Discussion} 'Discussion' Section exposes the importance of our approach to obtain accurate and interpretable models. Finally, Section~\ref{sec:Conclusion} 'Conclusion' highlights the main results.

\section{Background}
\label{sec:Background}

The proposed method embraces the explainability problem by exploiting the advantages and disadvantages of deep learning and radiomic workflow-based classification methods. Therefore, the section introduces general concepts related to XAI and particularizes deep architectures and radiomics. At the end of the section, the need and importance of the proposed method are justified.

\subsection{Explainable AI}

Numerous definitions of explainable AI are not fully standardized yet \cite{longo2024explainable}. For this article, it is essential to delineate certain concepts. A model is \textit{intrinsically interpretable} when its transparent structure allows an understanding of its decision process. For this reason, a transparent algorithm doesn't require explanation methods for its interpretation. From an algorithmic point of view, Logistic Regression, Decision Tree, and Naive Bayes are considered transparent, while Neural Networks-based methods are defined as \textit{black-boxes}. In the case of black-box algorithms, explanation methods are required for their introspection. An explanation can be global and local. A \textit{global explanation} focuses on explaining a model's behavior across its entire dataset. A \textit{local explanation} explains the prediction of a specific dataset sample (\textit{i.e.} patient). To an \textit{intelligible feature} it is possible to associate a human-understandable meaning. Radiomic features are considered intelligible. On the other hand, \textit{deep features} that are extracted using neural network-based architectures, lack interpretability (are not intelligible). We can call features extracted via convolutional neural networks \textit{CNN-derived}, and extracted via Vistion Transformer as \textit{ViT-derived}.

The combination of intelligible features and transparent algorithms makes the whole system intrinsically interpretable for two reasons: \textit{1)} it is possible to calculate the impact of the feature for model decision; \textit{2)} it is possible to correlate and validate the model's findings comparing it with the clinical literature. Local and global explanations are possible by employing black-box algorithms and intelligible features. These techniques are usually implemented to the already trained model and are known as \textit{post-hoc} algorithms. When not-intelligible features are used, model explainability becomes significantly more complex: model findings cannot be compared with clinical literature because each feature's meaning is unknown. A method is defined \textit{agnostic} when can be applied to any algorithm and architecture. For classification tasks, an XAI method can be \textit{class-dependent} when explanations are given for each class separately, conversely, they are defined as \textit{class-independent}.

\subsection{Explainability in Deep Architectures}

Convolutional-based neural networks have emerged as a standard for deep feature extraction in medical images. Deep learning has shown outstanding capabilities in diagnostic imaging across various diseases and modalities, highlighting its potential as a valuable clinical tool. Despite this promise, its adoption in clinical settings remains limited.  One of the main reasons is related to the lack of transparency and trust \cite{deVries2023explainable}. The idea of saliency maps, which is used to highlight the visual regions that are most important for the prediction, is the primary emphasis of these architectures' explainability \cite{Itti1998, mamalakis20233d}.
In \cite{Simonyan2013}, given an input image, the gradient computation of the class score is calculated to visualize the activation map for a particular class. Integrated gradients are proposed in \cite{Sundararajan2017} for a more robust explanation through activation maps, adding the constraint of respecting the \textit{'sensitivity'} and the \textit{'implementation invariance'} principles/axioms. In \cite{Zeiler2011} and \cite{Zeiler2014} a network using deconvolution is proposed to visualize the convolutional network while in \cite{Springenberg2014} the guided back-propagation is introduced to visualize the features learned from CNN. The most popular algorithms exploiting saliency belong to the category of 'class activation maps' (CAM) \cite{Zhou2016}. GradCAM \cite{Selvaraju2017} has been introduced after CAM, to overcome the limitation of being able to visualize only the last layer and only particular CNN architectures. Despite many other methods being proposed later, GradCAM is still one of the most popular. Lately, \textit{gradient-free} methods are also gaining wide popularity to overcome some limitations for computing gradients in very-deep architectures, and overcome the problem of anomalous behavior in incorrect prediction scenarios \cite{muhammad2020eigen}. 

Saliency maps are mainly used to provide local explanations and it is not clear how these tools can be used for global explanation. Furthermore, it is important to emphasize that criticisms have been raised about their use \cite{prinzi2023explainable}. According to \cite{oh2020deep, signoroni2021bs}, for instance, poorly localized and spatially blurred vision was discovered in certain instances. Furthermore, it has been demonstrated that various methods for computing saliency maps can yield contradictory outcomes \cite{prinzi2024yolo, Zhang2022overlooked}.

\subsection{Explainability in Radiomics}

Radiomics is gaining prominence for extracting features from radiological images. It operates on the premise that some pixel-wise patterns are hidden from the human eye \cite{gillies2016radiomics}. In contrast to deep features, radiomic features quantify certain properties like texture/pattern and statistical measurement by applying mathematical formulas to the images. For this reason, radiomic features are also called \textit{hand-crafted}, counterposed to the \textit{learned} extracted by neural networks. Radiomic flow has several advantages over deep extraction. 
Radiomic feature extraction does not require a large amount of data, as is the case with deep feature extraction \cite{wei2021radiomics}.
Moreover, shallow learning algorithms prove to be well-suited for classifying radiomic data, widely recognized as more appropriate for training with small datasets. The primary benefit of radiomic features is their inherent interpretability, as their meanings are well understood \cite{militello2023ct, prinzi2024micro}. Using shallow learning algorithms alongside these inherently interpretable features allows for both local and global explanations of the predictive models. \\

\begin{figure*}[t]
    \centering
        \includegraphics[width=\textwidth]{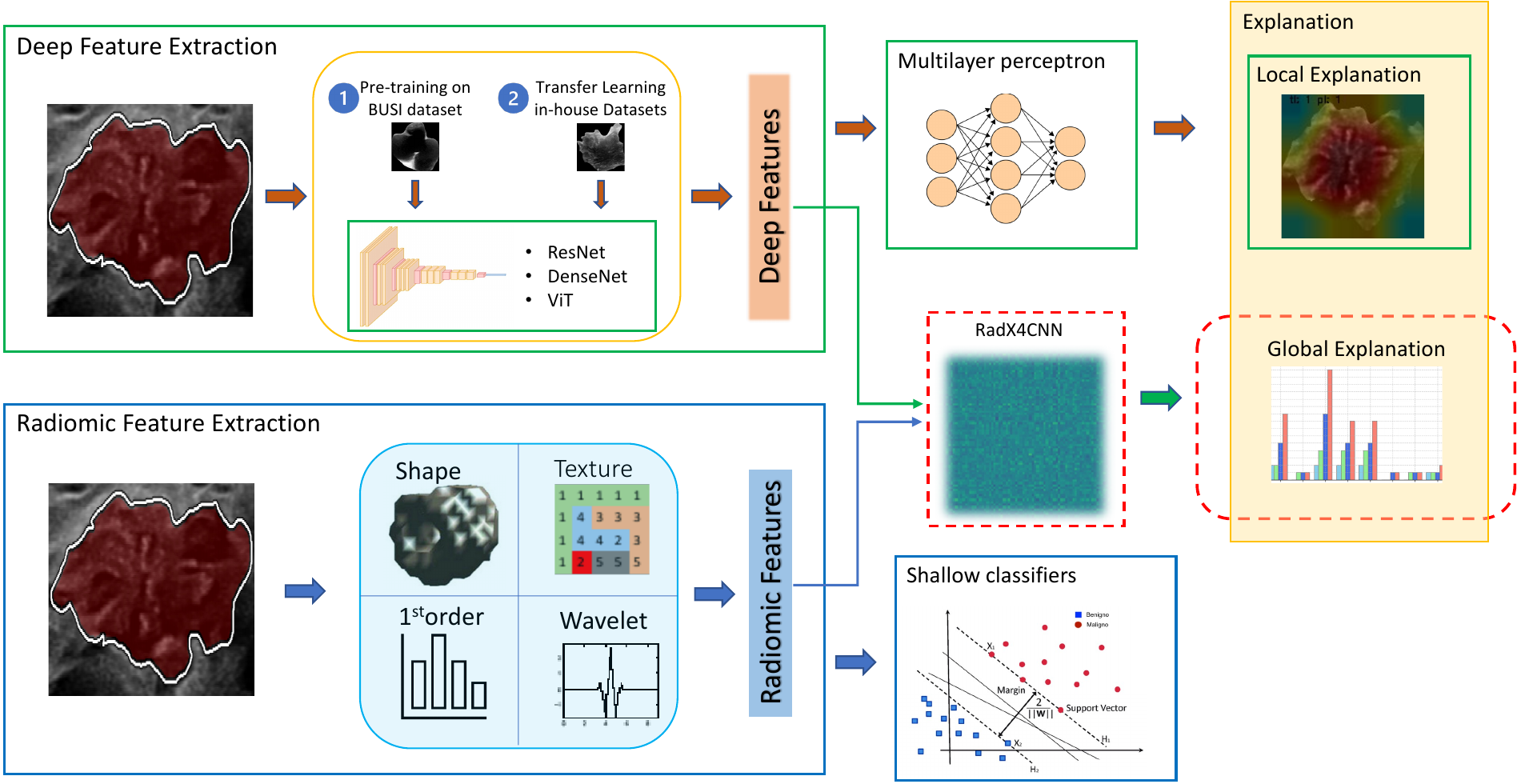}
            \caption{Overall workflow. Deep and Radiomic approaches were implemented for feature extraction and classification. Successively, radiomic and deep features were employed for \acrshort{MethodAcronym} implementation, providing a global explanation method.}
    \label{fig:workflow}
\end{figure*}

\subsection{Our Rationale}

Despite in some applications the use of transparent models can be sufficient \cite{Rudin2019}, the design of a transparent model can introduce a level of complexity \cite{chen2022explainable}: deep features tend to be more informative than interpretable radiomic features. In addition, the use of transparent algorithms (such as linear models) may fail when the relationships between the data are complex and nonlinear. As a consequence, the use of transparent-by-design models can lead to poor performance. For this reason, the eternal conflict between the explainability and accuracy trade-off comes into play.

The literature indicates the need for new methods for CNN explanation that deviate from those based on visualization maps \cite{papanastasiou2023attention}. Some attempts have been made by \cite{paul2019explaining, wang2019deep, chen2016automatic} although there is no explicit reference to XAI.  \cite{paul2019explaining} evaluated how radiomic model performance varies by replacing a correlated deep feature in the radiomic signature. However, regarding the performed interpretable feature extraction, the Image Biomarkers Standardization Initiative (IBSI) guidelines \cite{zwanenburg2020image} had not yet been proposed at that time. As a consequence, assessing the reproducibility of the study and the resulting impact of the study can be challenging. In addition, the authors do not use state-of-the-art deep extraction architectures such as ResNet, DenseNet, and ViT, whose depth is quite different from the VGG employed. Furthermore, they agree with an important limitation regarding the fact that a single slice was used to extract deep features in a volumetric image, whereas semantic information was generated from multiple slices. Above all, in contrast to our idea, they modify the input signature by replacing semantic features with deep ones, introducing a kind of constraint for model training. \cite{wang2019deep} attempts to exploit the radiologist's knowledge to qualitatively and visually correlate CNN-based feature maps with semantic features related to the investigated domain. However, these radiological features are operator-dependent, as is their correlation with the feature maps. As a result, their proof-of-concept is not generalizable to any context or domain. A similar idea was proposed in \cite{chen2016automatic}, to reduce the gap between deep features and various clinical semantic features. The model aims to yield semantic rating scores to support a deeper analysis for either clinical diagnosis or educational purposes. More recently, \cite{liu2023semantic} tried to train deep models to grade informative semantic features related to the analyzed task rather than directly predicting the disease malignancy.
Both papers lack formalism regarding the terminology related to explainable AI (\textit{i.e.}, the characteristics of the method in terms of global-local explanation, applicability, class dependencies, etc.) In general, in the attempt to make the predictive process transparent, the cited articles, employ semantic features subject to inter-operator variability or modify the training process, which could harm the established predictive performance of deep models.

In this paper, we present \acrshort{MethodAcronym}, a method that integrates deep features' predictive strengths with the interpretability of radiomic features. The method does not introduce any constraints in the training that can inhibit its accuracy. Considering the development of new explanation methods not solely based on visualization maps as a major challenge \cite{papanastasiou2023attention}, \acrshort{MethodAcronym} offers an innovative approach to explain deep architectures and provides a global explanation of deep networks, an aspect often overlooked in current research.

\section{Materials and Methods}
\label{sec:MaterialsMethods}

Figure~\ref{fig:workflow} shows the flow diagram of the proposed method. For the deep workflow ResNet50, DenseNet121, and ViT-B32 were pre-trained on the BUSI dataset. Then transfer learning was implemented using the two in-house datasets.
The extracted deep features were employed both for classification (via multilayer perceptron) and for explanation (local explanation via visualization maps). For the radiomic workflow, radiomic features were extracted, preprocessed, and used to train shallow classifiers. Radiomic and deep features were used as input for \acrshort{MethodAcronym} providing a global explanation.

\subsection{Datasets Description}

\paragraph{Breast Ultrasound Images Dataset (BUSI) -} The BUSI \cite{al2020dataset} comprises 780 ultrasound images obtained from 600 patients aged 25-75 years in 2018, alongside corresponding masks in PNG format. The images, containing the Focal Breast Lesions (FBLs) and measuring $500 \times 500$ pixels, are categorized into three classes: benign, malignant, and normal. Among these, 438 images present benign lesions, 211 are malignant, and 133 are classified as normal. Normal samples were not used in the study. The dataset exhibits class imbalance, with 67.49\% (438 images) benign and 32.51\% (211 images) malignant samples.

Table~\ref{tab:datasetDataAugmentation} provides the details concerning the number of images available of the BUSI dataset before and after the data augmentation, which was performed to increase the training set of the minority class: only the malignant class was augmented. Moreover, the table reports the dataset subdivision for training, validation, and test phases. After the initial 80:20 ratio between ‘Training/Validation’ and ‘Test', the ‘Training/Validation’ set was again divided by applying an 80:20 ratio to obtain the ‘Training’ and ‘Validation’ sets.

\begin{table*}
\centering
    \caption{Details concerning the number of images available of the BUSI dataset before and after the data augmentation, which was performed to increase the training set of the minority class. Moreover, the table reports the dataset subdivision for training, validation, and test phases. (*) indicates the number of images obtained after data augmentation.}
    \label{tab:datasetDataAugmentation}

    \begin{tabular}{l|c|c|c|c}
        \toprule
            \multirow{4}{*}{\textbf{Lesion Class}} & \multicolumn{3}{c|}{\textbf{Training/Validation (80\%)}} & \multirow{4}{*}{\textbf{Test (20\%)}} \\
        \cmidrule(lr){2-4}
            {} & \multicolumn{2}{c|}{\textbf{Training (80\%)}} & \multirow{2}{*}{\textbf{Validation (20\%)}} & {} \\
         \cmidrule(lr){2-3}
         {} & \textbf{before data augmentation} & \textbf{after data augmentation} & {} & {}\\
        \midrule
            {Benign} & {280} & {280} & {70} & {88} \\
            {Malignant} & {135}  & {270 $^{(*)}$} & {34} & {42} \\
        \midrule
            \textbf{Total} & \textbf{415} & \textbf{550} & \textbf{104} & \textbf{130} \\
        \bottomrule
    \end{tabular}
\end{table*}

\begin{table}
\centering
    \caption{Description of the in-house collected dataset, coming from Palermo and Cefalù hospitals.}\label{tab:datasetCharacteristics}
    \begin{tabular}{l|cc}
        \toprule
            \textbf{Characteristic} & \textbf{Palermo Dataset} & \textbf{Cefalù Dataset} \\
        \midrule
            {image size} & {845 $\times$ 600} & {845 $\times$ 600}  \\
            {focal breast lesions} & {237} & {115}  \\
            {benign lesion} & {132} & {70}  \\
            {malignant lesion} & {105} & {45}  \\
            {size range [\textit{mm}]} & {4-90} & {3-50}  \\
            {mean size [\textit{mm}]} & {15.12 $\pm$ 9.44} & {14.64 $\pm$ 8.61}  \\
            {age range [\textit{years}]} & {17–88} & {23-89}  \\
            {mean age [\textit{years}]} & {53.15 $\pm$ 15.00} & {51.61 $\pm$ 14.56}  \\
        \bottomrule
    \end{tabular}
\end{table}

\begin{table}
\centering
    \caption{Details on proprietary datasets: splits employed for model training and test.}
    \label{tab:proprietaryDatasets}

    \begin{tabular}{l|ccc|c}
        \toprule
            \multirow{2}{*}{\textbf{Lesion Class}} & \multicolumn{3}{c|}{\textbf{Palermo Dataset}} & \textbf{Cefalù Dataset } \\
        \cmidrule(lr){2-5}
            {} & \textbf{Total} & \textbf{Training} & \textbf{Test} & \textbf{Test} \\
        \midrule
            {Benign} & {132} & {106} & {26} & {70} \\
            {Malignant} & {105} & {84} & {21} & {45} \\
        \midrule
            \textbf{Total} & \textbf{237} & \textbf{190} & \textbf{47} & \textbf{115} \\
        \bottomrule
    \end{tabular}
\end{table}

\paragraph{Palermo and  Cefalù Datasets -} Two \textit{in-house} datasets were collected on two centers, referred to as \textit{Palermo} and \textit{Cefalù} datasets. Considering the Palermo site, 237 breast cancer patients were enrolled in 2021 at Policlinico Universitario “P. Giaccone”. This dataset was employed for model tuning and internal test. Instead, considering the Cefalù site, 115 breast cancer patients were enrolled in 2022 at Fondazione Istituto "G. Giglio" Breast Unit. This dataset was used only for model external validation. Table~\ref{tab:datasetCharacteristics} provides a detailed explanation of the in-house datasets.
FBL images of both datasets were acquired through the B-mode ultrasound modality. Two expert breast radiologists (one from each institution with over 30 years of expertise in breast imaging) used two identical ultrasound RS85 (Samsung Medison, Co. Ltd.) devices, each with a 3–12 MHz linear transducer. Acquired FBLs were labeled using the ultrasound BI-RADS criteria \cite{sickles2014should}, and successively the images were segmented. The adopted inclusion criteria for our breast ultrasound in-house datasets are discussed here \cite{bartolotta2021s}. 
Patients had not received any intervention or surgery on lesions before the ultrasound examination. Lack of biopsy or irregular follow-up were the exclusion criteria \cite{bartolotta2024artificial}.  
Table~\ref{tab:proprietaryDatasets} shows the split of the proprietary datasets used for training and testing. A cross-validation was implemented using the training set.

\subsection{Deep Architecture Training}

\begin{figure*}[t]
    \centering
    \includegraphics[width=\textwidth]{{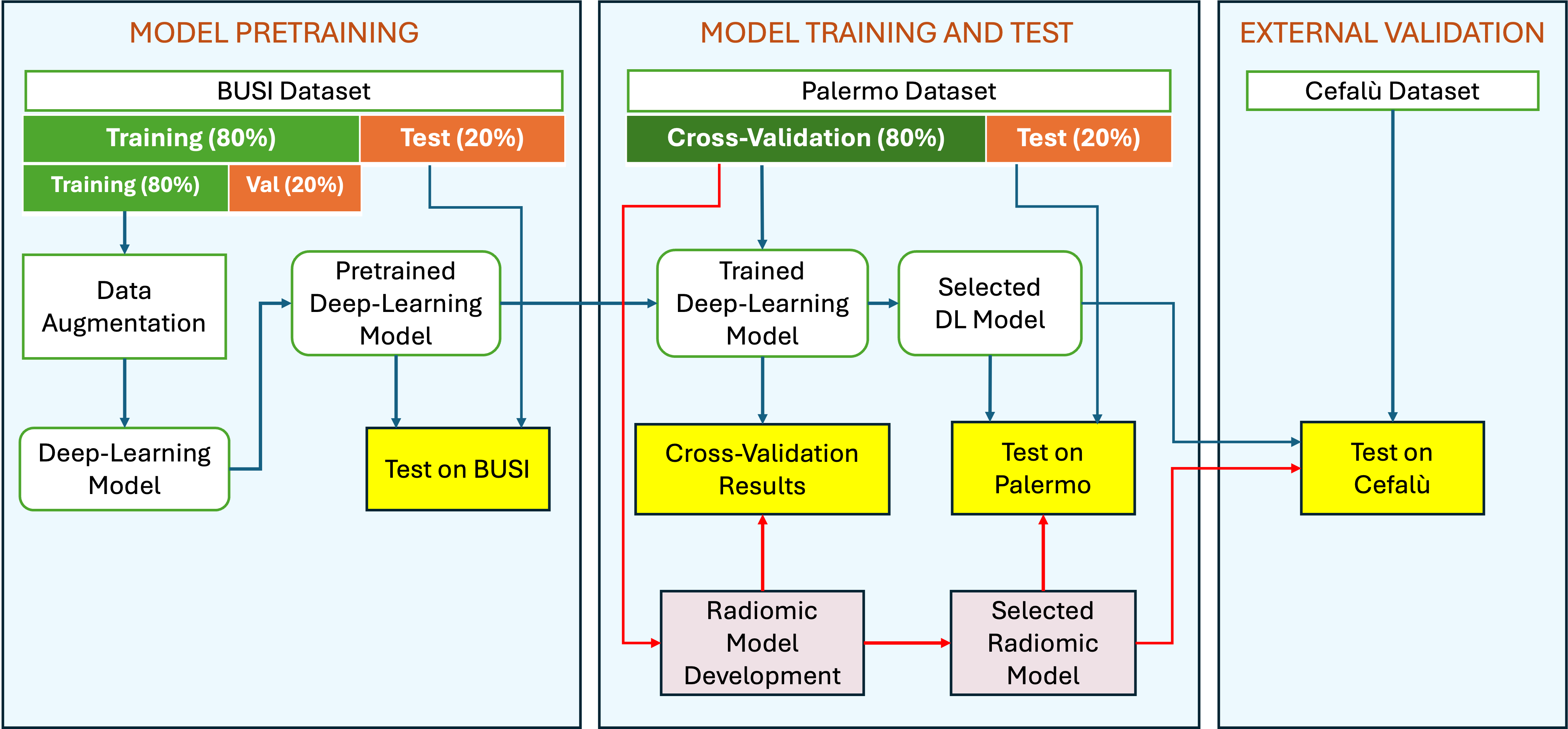}}
    \caption{Diagram depicting the details about the training, the testing, and the external validation related to deep and radiomic workflow.}
    \label{fig:workflowRadiomicsTraining}
\end{figure*}

This section includes a description of the workflow used to train deep architectures. In particular, preprocessing strategies, data preparation for training and test, and the implemented architectures are discussed. In addition, state-of-the-art methods for deep features explanation were exposed as well as our novel \acrshort{MethodAcronym} method. The diagram reported in Figure~\ref{fig:workflowRadiomicsTraining} depicts the details about the training, the testing, and the external validation related to deep workflow.

\paragraph{Image preprocessing -} Tumor patches were extracted from the whole image using the available masks. All patches were resized to size $128 \times 128$, specifically: \textit{i) }images with a size larger than $128 \times 128$ were subsampled, and \textit{ii)} zero-padding was applied to images with a smaller size. Images were normalized before model training.

\paragraph{Training and test protocol -} Using an 80:20 ratio, the BUSI dataset was split into training and test sets. With the same ratio, the training set was divided once more into training and validation sets.
Whereas having unbalanced datasets can lead to models affected by bias \cite{hasib2020survey}, the training set was balanced by considering random flip, rotation, contrast enhancement, translation, and zoom. The purpose of the BUSI dataset, in light of the recent criticism \cite{pawlowska2023dib}, was to produce an optimized pre-trained model solely.
Regarding the Palermo dataset, 20\% was used as an independent test set, while the remaining 80\% was used to implement a 5-fold cross-validation (CV). The Cefalù dataset was used only for external validation. The best model in terms of accuracy within the CV procedure was selected for the Palermo internal test and the Cefalù external validation. 

\paragraph{Saliency Maps Computation -} Saliency maps highlight regions of an input image that most contribute to the prediction. Considering a classification task with $c$ classes and a convolutional neural network with $A^k$ feature map activations, Grad-CAM \cite{Selvaraju2017} is formalized as:

\begin{equation} \label{equ:gradCAM1}
    L^{c}_{Grad-CAM} = ReLU(\sum_k \alpha^{c}_{k} A^k_l)
\end{equation}

where: 

\begin{equation} \label{equ:gradCAM2}
    \alpha^{c}_{k} = \frac{1}{z} \sum_i \sum_j \frac{\partial y^c}{\partial A^{k}_{ij}}
\end{equation}

Here $\frac{1}{z} \sum_i \sum_j$ represents the global average pooling operator over the width and height dimensions (indexed by $i$ and $j$
respectively) and $\frac{\partial y^c}{\partial A^{k}_{ij}}$ gradients via backpropagation. The weights $\alpha^{c}_{k}$ reflect a partial linearization and quantify feature map $k$ importance for a given target class $c$. Grad-CAM's dependence on class $c$ makes it a class discriminative method.
Although Grad-CAM appears as one of the most widely used approaches, has two main limitations related to the gradient computation: \textit{i)} saturation: The zero-gradient area of the ReLU function or the saturation problem for the Sigmoid function may cause the gradient to vanish; and \textit{ii)} false confidence: when compared to a zero baseline, activation maps with larger weights exhibit smaller contributions to the network's output \cite{wang2020score}. Furthermore, backpropagating any quantity has extra computing overhead and relies on the assumption that classifiers generated accurate prediction; in the event that an incorrect decision is made, all of the aforementioned techniques will result in inaccurate or distorted representations \cite{muhammad2020eigen}. \\

Score-CAM introduced the Channel-wise Increase of Confidence (CIC), in contrast to Grad-CAM \cite{Selvaraju2017}, which uses the gradient information coming into the last convolutional layer to reflect the relevance of each activation map \cite{wang2020score}.

\begin{equation} \label{equ:scoreCAM1}
    L^{c}_{Score-CAM} = ReLU(\sum_k \alpha^{c}_{k} A^k_l)
\end{equation}

with: 

\begin{equation} \label{equ:scoreCAM2}
    \alpha^{c}_{k} = C(A^k_l)
\end{equation}

where the CIC, used to gauge each activation map's significance, is shown by the symbol $C(\cdot)$.

EigenCAM is a gradient-free method and uses the principal components from the extracted feature maps \cite{muhammad2020eigen}. For this reason, it can overcome the problem of distorted displays in case of incorrect predictions. Let $W_{L=k}$ represent the combined weight matrix of the first $k$ layers of size $(m, n)$, and let $I$ represent the input image of size $(i \times j)$, $I \in \mathbb{R}^{i \times j}$. The image $I$ projected onto the last convolution layer $L=k$ is the class-activated output, and it is provided by $O_{L=k} = W^{T}_{L=k} I$. The principal components of $O_{L=k}$ can be computed by factorizing $O_{L=k}$ using singular value decomposition, which yields $O_{L=K} = U \Sigma V^T$. $\Sigma$ is a diagonal matrix of size $M \times N$ with singular values along the diagonal, and $V$ are the left singular vectors. $U$ is an orthogonal matrix of size $M \times M$, and the left singular vectors are found in the column $U$. The projection of $O_{L=k}$ on the first eigenvector yields the class activation map:

\begin{equation} \label{equ:eigenCAM1}
    L_{EigenCAM} = O_{L=K} V_1 
\end{equation}

where the first eigenvector in the V matrix is $V_1$. Eigen-CAM is class-independent, \textit{e.g.} the result of saliency maps is not tied to any class.\\

The three methods (\textit{e.g.} Grad-CAM, Score-CAM, and Eigen-CAM) were implemented to compare the different explanations provided and, thus, highlight existing gaps and the need for alternative explanation techniques.

\subsection{Radiomic Models Training}

Radiomic workflow includes ROI segmentation, feature extraction, feature preprocessing, selection, and model training. The diagram reported in Figure~\ref{fig:workflowRadiomicsTraining}, already introduced in the previous section, depicts the details about the training, the testing, and the external validation related to radiomic workflow.

\paragraph{ROI Segmentation -} Radiologists used S-DetectTM to perform ROI segmentation from each B-mode image of the breast lesion. Installed on the RS85 ultrasound machine, this software is licensed for clinical usage and is available for purchase. 

\paragraph{Feature Extraction -} PyRadiomics \cite{griethuysen2017computational} is a toolkit that complies with the IBSI \cite{zwanenburg2020image}, and was used to extract a total of 474 radiomic features. By standardizing the radiomic analysis procedure, IBSI increases workflow reproducibility.
Extracted features belong to the following categories: shape 2D; first-order intensity histogram statistics; Gray Level Co-occurrence Matrix features (GLCM); Gray Level Run Length Matrix (GLRLM); Gray Level Size Zone Matrix (GLSZM); Gray Level Dependence Matrix (GLDM); Neighboring Gray Tone Difference Matrix (NGTDM). Features were extracted discretizing images to 255 gray levels. Furthermore, considering the high productivity of wavelet-derived radiomic features in comparison to the original ones \cite{prinzi2023impact}, non-shape features were extracted also considering the Haar wavelet transform.

\paragraph{Features Preprocessing and Selection -} Non-redundant and informative features were selected \cite{militello2022robustness, papanikolaou2020develop} through:  \textit{i)} near-zero variance analysis to eliminate features with poor information content (variance cutoff $\sigma^2 =0.005$); \textit{ii)} correlation analysis to exclude strongly correlated features (the Spearman correlation with a 0.9 threshold value was evaluated). Eventually, the Sequential Feature Selector (SFS) algorithm was employed in conjunction with the Random Forest (RF), XGBoost (XGB), and Support Vector Machine (SVM) classifiers. Accuracy was used as a metric to maximize.

\paragraph{Predictive Model Setup -} This study employed three distinct shallow classifiers: RF, XGB, and SVM. These algorithms have proven suitable for tabular data in small dataset scenarios \cite{prinzi2022dcemri}. RF and XGB were trained using 100 estimators. For XGB a learning rate of 0.3 and gain as the importance type were employed. SVM was trained using RBF kernel and features were standardized. A stratified 10-fold Cross-Validation approach that was repeated 20 times was taken into consideration. To perform the test phase, the model with the highest accuracy after 20 repetitions of a 10-fold cross-validation process was selected.

\subsection{Deep Features Explanation by means of Radiomics}

\acrshort{MethodAcronym} aims to explain the deep features extracted through the intrinsic meaning of radiomic features. Let $ f_r = (r_1, r_2, ..., r_{a}) $ denote the radiomic features, and $ f_d = (d_1, d_2, ..., d_{b}) $ denote the deep features, where $ a $ and $ b $ are the number of radiomic and deep features, respectively. The method assumes the use of a correlation index, and Sperman's test was employed in this work.
The Spearman correlation coefficient between $ f_r $ and $ f_d $ is given by:

\begin{equation}
    \rho_{i,j} = \frac{{\text{cov}(r_i,d_j)}}{{\sigma_r \sigma_d}} \forall i \in [1, a] \text{, } j \in [1, b] \text{, with} i,j \in \mathbb{N}
\end{equation}

where: $ \text{cov}(r_i,d_j) $ is the covariance between $ r_i $ and $ d_j $, $ \sigma_r $ is the standard deviation of $ r_i $, $ \sigma_d $ is the standard deviation of $ d_j $.

To explain the meaning of $ f_d $ exploiting the correlation with the features $ f_r $ using a thresholds $M$, a threshold function $ T(\rho, \text{M}) $ is defined such that:

\begin{equation}
    T(\rho, \text{M}) = \begin{cases} 1 & \text{if } \rho \geq \text{M} \\ 0 & \text{otherwise} \end{cases} 
\end{equation}

where: $ \rho $ is the Spearman correlation coefficient between $ f_r $ and $ f_d $, $M$ is the threshold value. $M$ represents a method hyperparameter to be determined according to the specific case study. By applying different threshold values, it is possible to determine which features in $ f_d $ are correlated with radiomic features $ f_r $. It assumed that correlated radiomic and deep features describe the same phenomenon (\textit{e.g.} features have a similar meaning) \cite{paul2019explaining, wang2019deep}. Below are the main characteristics of the proposed method:

\paragraph{Global Explanation -} Our method implements the notion of global explanation, diverging from approaches reliant on computing saliency maps, which constitute a form of local explanation. Global explanation is a key aspect of clinical model validation as it facilitates a comparative analysis of model findings against the existing clinical literature.

\paragraph{Model-agnostic -} This method is agnostic, making it universally applicable across architectures and application contexts. It merely necessitates the extraction of feature vectors $ f_d $ and $ f_r $, rendering it adaptable to any model and application without constraints. In agreement with the other saliency maps methods, \acrshort{MethodAcronym} need the vectorized version $ f_d $ of features maps $A^{k}_{ij}$, such as Grad-CAM, Score-CAM and $W_{L=k}$ Eigen-CAM.

\paragraph{Class-Independent -} \acrshort{MethodAcronym} explains deep features without class specificity, such as Eigen-CAM. Conversely, it contrasts with Grad-CAM and Score-CAM, which are class-dependent considering the computation of $\alpha^{c}_{k}$.

\paragraph{Ease of Implementation -} It requires only the extraction of $ f_d $, which is inherently part of any training process, along with the extraction of radiomic features $ f_r $ and the subsequent calculation of their correlation. The simplicity of implementation is the basis of several widely adopted and popular techniques \cite{muhammad2020eigen, Zeiler2014}.

\begin{table*}[htbp]
    \centering
    \caption{Cross-validation performance comparison of VGG, ResNet, ViT, and the radiomic model on the Palermo training dataset.}
    \label{tab:performance_val}
    \begin{tabular}{l|cccccc}
        \toprule
            \textbf{Model} & \textbf{Acc} & \textbf{AUROC} & \textbf{Sens} & \textbf{Spec} & \textbf{PPV} & \textbf{NPV} \\
        \midrule
            ResNet50 & $93.62 \pm 9.19$ & $0.95 \pm 0.09$ & $91.20 \pm 14.83$ & $96.26 \pm 4.6$ & $95.23 \pm 5.91$ & $93.01 \pm 12.00$ \\
            DenseNet121 & $93.17 \pm 13.66$ & $0.94 \pm 0.12$ & $93.33 \pm 13.33$ & $93.00 \pm 14.00$ & $93.33 \pm 13.33$ & $93.00 \pm 14.00$ \\
            ViT-B32 & $93.65 \pm 11.5$ & $0.97 \pm 0.6$ & $92.16 \pm 12.95$ & $95.00 \pm 10.00$ & $94.74 \pm 10.53$ & $92.80 \pm 12.42$ \\
            Rad + XGB & $61.49 \pm 9.88$ & $0.65 \pm 0.10$ & $53.63 \pm 16.79$ & $67.53 \pm 14.13$ & $56.63 \pm 13.19$ & $66.05 \pm 9.59$ \\
            Rad + RF & $59.93 \pm 9.19$ & $0.60 \pm 0.11$ & $39.61 \pm 14.31$ & $75.35 \pm 13.01$ & $56.66 \pm 17.76$ & $62.10 \pm 7.34$ \\
            Rad + SVM & $65.05 \pm 8.31$ & $0.62 \pm 0.10$ & $55.83 \pm 15.05$ & $72.09 \pm 12.13$ & $61.52 \pm 10.06$ & $68.70 \pm 8.60$ \\
        \bottomrule
    \end{tabular}
\end{table*}

\begin{table*}[htbp]
    \centering
    \caption{Internal test and external validation performance using ResNet, DenseNet, ViT, and radiomic models on the Palermo and Cefalù test datasets.}
    \label{tab:performance_test}
    \begin{tabular}{l|cccccc|cccccc}
        \toprule
            {} & \multicolumn{6}{c|}{\textbf{Palermo Hospital Dataset}} & \multicolumn{6}{c}{\textbf{Cefalù Hospital Dataset}} \\
        \cmidrule(lr){2-7} \cmidrule(lr){8-13}
            \textbf{Model} & \textbf{Acc} & \textbf{AUROC} & \textbf{Sens} & \textbf{Spec} & \textbf{PPV} & \textbf{NPV} & \textbf{Acc} & \textbf{AUROC} & \textbf{Sens} & \textbf{Spec} & \textbf{PPV} & \textbf{NPV} \\
        \midrule
            ResNet50 & $ 72.22 $ & $ 0.733 $ & $ 66.67 $ & $ 76.19 $ & $ 66.67 $ & $ 76.19 $ & $ 73.04 $ & $ 0.757 $ & $ 77.78 $ & $ 70.00 $ & $ 62.5 $ & $ 83.05 $ \\
            DenseNet121 & $ 77.78 $ & $ 0.784 $ & $ 73.33 $ & $ 80.95 $ & $ 73.33 $ & $ 80.95 $ & $ 70.44 $ & $ 0.755 $ & $ 68.89 $ & $ 71.43 $ & $ 60.78 $ & $ 78.13 $\\
            ViT-B32 & $ 83.33 $ & $ 0.848 $ & $ 93.33 $ & $ 76.19 $ & $ 73.68 $ & $ 94.12 $ & $ 62.61 $ & $ 0.742 $ & $ 77.78 $ & $ 52.86 $ & $ 51.47 $ & $ 78.72 $\\
            Rad + XGB & $ 47.22 $ & $ 0.43 $ & $ 46.15 $ & $ 47.82 $ & $ 33.33 $ & $ 47.22 $ & $ 61.11 $ & $ 00.60 $ & $ 60.86 $ & $ 61.35 $ & $ 62.89 $ & $ 60.86 $\\
            Rad + RF & $ 58.33 $ & $ 0.61 $ & $ 61.15 $ & $ 56.52 $ & $ 44.44 $ & $ 58.33 $ & $ 51.12 $ & $ 0.55 $ & $ 54.78 $ & $ 53.47 $ & $ 56.25 $ & $ 54.78 $\\
            Rad + SVM & $ 58.33 $ & $ 0.58 $ & $ 53.84 $ & $ 60.86 $ & $ 43.75 $ & $ 58.33 $ & $ 58.17 $ & $ 0.62 $ & $ 58.26 $ & $ 58.08 $ & $ 60.09 $ & $ 58.26 $\\
        \bottomrule
    \end{tabular}
\end{table*}

\section{Experimental Results}
\label{sec:ExperimentalResults}

Deep models were trained using Adam optimizer and 8 as batch size. For ResNet50 and DenseNet121 $10^{-4}$ was used as the learning rate, and $10^{-5}$ for ViT. A rate of 0.5 was used for dropout layers.
The BUSI dataset was employed to obtain an initial pre-trained model, and 80\% of the Palermo dataset was used for fine-tuning. Then, the remaining 20\% of the Palermo dataset was used as an internal test, and the whole Cefalù dataset for external validation.

\subsection{Deep and Radiomic Models Performance}

Table~\ref{tab:performance_val} provides a comparison between radiomic and deep models considering the Palermo training dataset. ResNet50, DenseNet121, and ViT-B32, demonstrate remarkable performance compared to radiomic models.
ResNet achieves an impressive accuracy of $93.62\% \pm 9.19$ and AUROC of $0.95 \pm 0.09$, showcasing its robustness in classification tasks. Similarly, DenseNet and ViT exhibit high accuracies and AUROC scores, proving the dominance of deep learning-based approaches over radiomic models. In contrast, models incorporating radiomic features coupled with traditional machine learning algorithms such as XGB, RF, and SVM perform poorly compared to deep architectures.

Internal test and external validation performance are shown in Table~\ref{tab:performance_test}, in which again the superiority of deep models over radiomic-based approaches is shown. In the Palermo dataset, ResNet achieves an accuracy of $72.22\%$ and AUROC of $0.733$, while DenseNet and ViT achieve even higher accuracies and AUROC scores. These results highlight the effectiveness of deep learning methodologies in extracting meaningful features from the datasets, leading to improved diagnostic performance. On the Cefalù dataset, a similar trend is observed, with deep models outperforming radiomic-based approaches. In particular, ResNet proved the best generalization capability on the external validation, showing an accuracy of $73.04\%$ and a good balance between sensitivity and specificity. It is important to emphasize that although the performance of ResNet on the internal test is much lower than ViT, on the external validation ResNet maintains its performance, as opposed to ViT which achieves a heavy degradation. For this reason, we consider ResNet the best-trained model, and further considerations on explainability are provided for ResNet.

In contrast, models incorporating radiomic features combined with traditional machine learning algorithms consistently demonstrate inferior performance compared to deep learning models on both internal test and external validation datasets. These results underscore the limitations of radiomic-based approaches in capturing complex patterns and features present in medical images, particularly when compared to the feature extraction capabilities of deep learning architectures.

\subsection{Traditional CNN Explainability}

\begin{figure*}[htb]
  \centering

  \sbox\PictureBox\ 
  \fboxsep=0pt%
  \raisebox{0.04\linewidth}{\fbox{%
      \parbox[b][\heightof{\usebox\PictureBox}-\fboxrule-\fboxrule][t]{0.169\linewidth}{\;}%
      }}
  \sbox\PictureBox\ 
  \fboxsep=0pt%
  \raisebox{0.04\linewidth}{\fbox{%
      \parbox[b][\heightof{\usebox\PictureBox}-\fboxrule-\fboxrule][t]{0.2\linewidth}{\centering \textbf{(a) original image}}%
      }}
        \sbox\PictureBox\ 
  \fboxsep=0pt%
  \raisebox{0.04\linewidth}{\fbox{%
      \parbox[b][\heightof{\usebox\PictureBox}-\fboxrule-\fboxrule][t]{0.2\linewidth}{\centering \textbf{(b) GradCAM}}%
      }}
        \sbox\PictureBox\ 
  \fboxsep=0pt%
  \raisebox{0.04\linewidth}{\fbox{%
      \parbox[b][\heightof{\usebox\PictureBox}-\fboxrule-\fboxrule][t]{0.2\linewidth}{\centering \textbf{(c) EigenCAM}}%
      }}
  \sbox\PictureBox\ 
  \fboxsep=0pt%
  \raisebox{0.04\linewidth}{\fbox{%
      \parbox[b][\heightof{\usebox\PictureBox}-\fboxrule-\fboxrule][t]{0.2\linewidth}{\centering \textbf{(d) ScoreCAM}}%
      }}
  
  \sbox\PictureBox\ 
  \fboxsep=0pt%
  \raisebox{0.2\linewidth}{\fbox{%
      \parbox[b][\heightof{\usebox\PictureBox}-\fboxrule-\fboxrule][t]{0.17\linewidth}{\textbf{Row 1: \\ benign lesion (correctly classified)}}%
      }}
  \usebox\PictureBox{%
    \includegraphics[width=0.2\linewidth,height=0.2\linewidth]{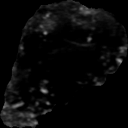}        \includegraphics[width=0.2\linewidth,height=0.2\linewidth]{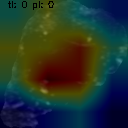}
    \includegraphics[width=0.2\linewidth,height=0.2\linewidth]{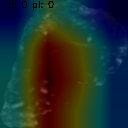}
    \includegraphics[width=0.2\linewidth,height=0.2\linewidth]{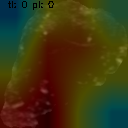}}

  \raisebox{0.2\linewidth}{\fbox{%
      \parbox[b][\heightof{\usebox\PictureBox}-\fboxrule-\fboxrule][t]{0.17\linewidth}{\textbf{Row 2: \\ malignant lesion (correctly classified)}}%
      }}
  \usebox\PictureBox{%
    \includegraphics[width=0.2\linewidth,height=0.2\linewidth]{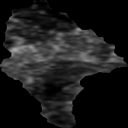}        \includegraphics[width=0.2\linewidth,height=0.2\linewidth]{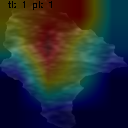}
    \includegraphics[width=0.2\linewidth,height=0.2\linewidth]{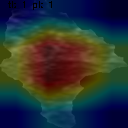}
    \includegraphics[width=0.2\linewidth,height=0.2\linewidth]{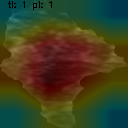}}

  \raisebox{0.2\linewidth}{\fbox{%
      \parbox[b][\heightof{\usebox\PictureBox}-\fboxrule-\fboxrule][t]{0.17\linewidth}{\textbf{Row 3: \\ benign lesion (wrongly classified)}}%
      }}
  \usebox\PictureBox{%
    \includegraphics[width=0.2\linewidth,height=0.2\linewidth]{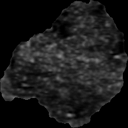}        \includegraphics[width=0.2\linewidth,height=0.2\linewidth]{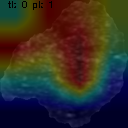}
    \includegraphics[width=0.2\linewidth,height=0.2\linewidth]{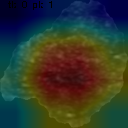}
    \includegraphics[width=0.2\linewidth,height=0.2\linewidth]{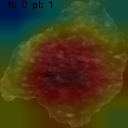}}

  \raisebox{0.2\linewidth}{\fbox{%
      \parbox[b][\heightof{\usebox\PictureBox}-\fboxrule-\fboxrule][t]{0.17\linewidth}{\textbf{Row 4: \\ malignant lesion (wrongly classified)}}%
      }}
  \usebox\PictureBox{%
    \includegraphics[width=0.2\linewidth,height=0.2\linewidth]{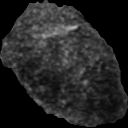}        \includegraphics[width=0.2\linewidth,height=0.2\linewidth]{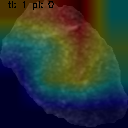}
    \includegraphics[width=0.2\linewidth,height=0.2\linewidth]{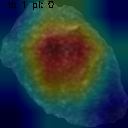}
    \includegraphics[width=0.2\linewidth,height=0.2\linewidth]{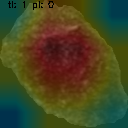}}

    \caption{(a) original ultrasound image; (b) ultrasound image with overlapped the GradCAM saliency map; (c) ultrasound image with overlapped the EigenCAM saliency map; (d) ultrasound image with overlapped the ScoreCAM saliency map.}
  \label{fig:saliencyMaps}
\end{figure*}

As the most accurate model in the external validation, ResNet was selected to analyze the traditional explainability methods based on saliency maps. By comparing the saliency maps obtained with GradCAM, EigenCAM, and ScoreCAM methods, it is possible to observe conflicting results among these interpretability techniques. Figure~\ref{fig:saliencyMaps} shows some examples. Despite ResNet's robust performance, the interpretations provided by these methods often diverged, presenting a challenge in understanding the model decision. This discrepancy raises questions about the reliability and consistency of these interpretability methods, urging a closer examination of their underlying mechanisms. Similar inconsistencies are recently achieved in \cite{cerekci2024quantitative, Zhang2022overlooked}. 
In particular, Rows 1-2 show two examples of correctly classified lesions, while Rows 3-4 show two examples of wrongly classified lesions. Some explanations predominantly focused on background regions, even when the model's prediction was correct. Other concerns arise in the case of wrong prediction. Considering Grad-CAM, this effect is due to the dependence of the gradient calculation of $\frac{\partial y^c}{\partial A^{k}_{ij}}$ (see Equation~\ref{equ:gradCAM2}) that is a function of the prediction $y^c$, assumed correctly predicted the class $c$. For Score-CAM this problem is mitigated by CIC which weights the importance of each activation map. For Eigen-CAM decomposition does not depend on the predicted class and gradient computation. This phenomenon underlines the complexity of interpreting neural network decisions and highlights the need for newer interpretability techniques capable of overcoming these problems and providing clearer and more actionable insights.

\subsection{Explaining Deep Features by means of Radiomics in Ultrasound Breast Dataset}
\label{ch:results_method}

\begin{figure*}[t]
    \centering
        \includegraphics[width=\textwidth]{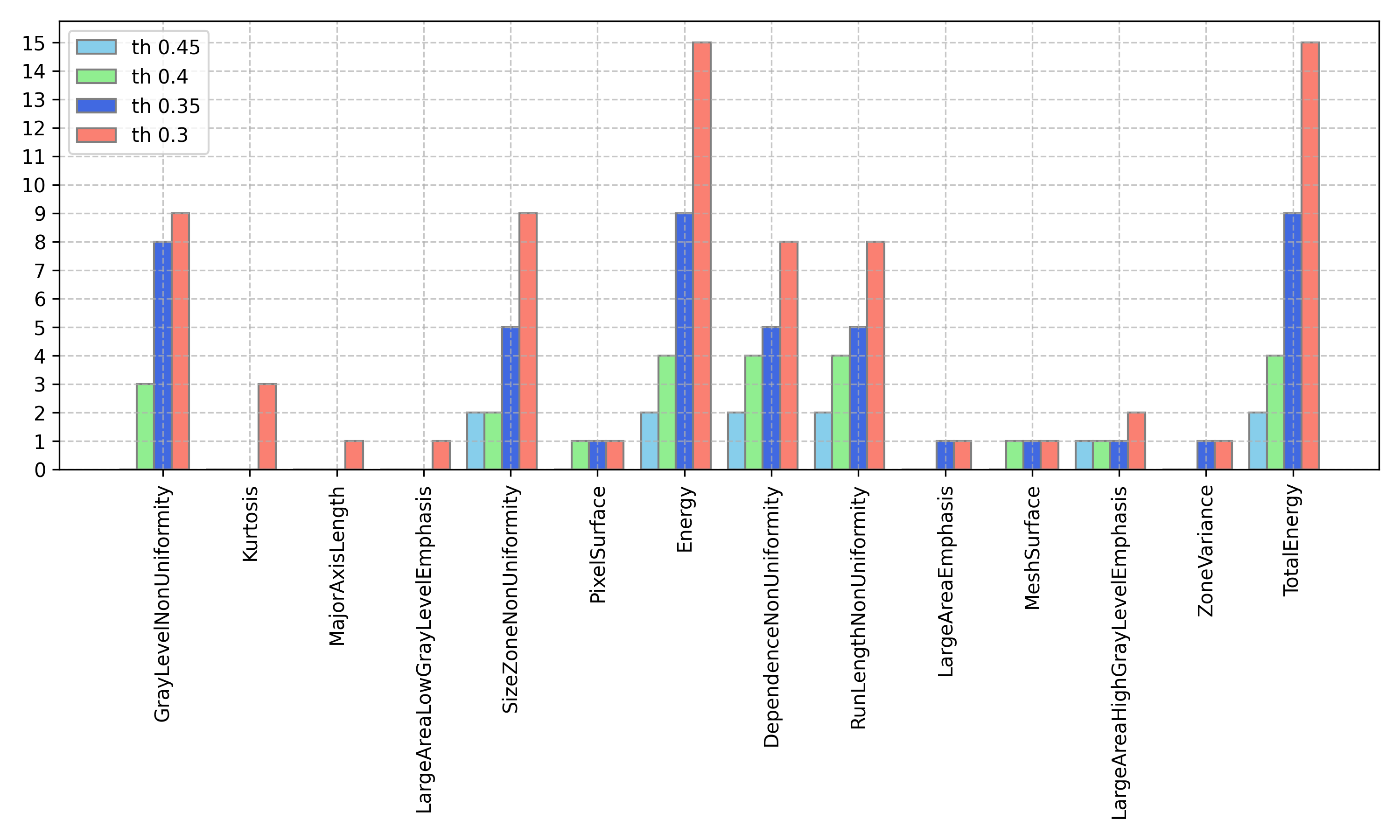}
            \caption{Global explanation of CNN-derived features obtained using \acrshort{MethodAcronym}. Each bar represents the number of CNN-derived features showing a correlation with radiomic features. The Y-axis reports the number of deep features correlated with the radiomic feature specified in the X-axis.}
        \label{fig:xai_corr}
\end{figure*}

Figure~\ref{fig:xai_corr} depicts the \acrshort{MethodAcronym} results, utilizing four distinct threshold values $\rho = [0.30, 0.35, 0.40, 0.45]$. The graph exclusively illustrates correlations with deep features derived from convolutional networks, as no correlations were identified with features extracted from VITs. Regarding the CNN-derived features (extracted from ResNet), our analysis reveals that \emph{Energy}, and \emph{TotalEnergy} emerge as the most prominently correlated features, with 15 correlations observed at a threshold of 0.3, 9 at 0.35, 4 at 0.4, and 2 at 0.45. Additionally, a high correlation (with $\rho = 0.45$) was established with \emph{SizeZoneNonUniformity}, \emph{DependenceNonUniformity}, and \emph{RunLengthNonUniformity}. These 5 radiomic features (\textit{i.e.} \emph{Energy}, \emph{TotalEnergy}, \emph{SizeZoneNonUniformity}, \emph{DependenceNonUniformity}, and \emph{RunLengthNonUniformity}) collect all the correlations, both those with the original (unfiltered) version of the feature and those derived after filtering  (considering the wavelets, with the 4 decompositions).

These correlations pave the way for pertinent clinical discussions, with implications already resonating within clinical practice. Further exploration of these clinical considerations is undertaken in Section~\ref{ch:clinical_validation}.

\section{Discussion}
\label{sec:Discussion}

The concept of XAI is attracting much interest \cite{holzinger2019causability, Stogiannos2023analysis} and is playing a key role in the implementation and deployment of eXplainable Clinical Decision Support Systems (X-CDSSs). In this perspective, our work introduces several novelties. \acrshort{MethodAcronym} does not impose any methodological constraints that can harm the model performance to introduce explainability. In addition, it is agnostic, \textit{e.g.} it is possible to apply it to any CNN-based architecture. Our findings allowed us \textit{i)} to find a substantial difference between radiomic, CNN-derived, and ViT-derived features, in terms of accuracy and explainability; \textit{ii)} to analyze some concerns related to the field of XAI, which our method attempts to mitigate; \textit{iii)} to highlight important clinical aspects in the context of breast cancer that are consistent with the clinical literature.

\subsection{Comparing ViT-derived, CNN-derived and Radiomic Features}

\acrshort{MethodAcronym} enabled an analysis of the role of radiomic, CNN-derived, and ViT-derived features, in terms of accuracy, explainability, and intrinsic meaning of deep features.

\paragraph{In terms of performance -} The ability of deep architectures to extract high and low-level features makes deep features much more informative than radiomic ones. As a consequence, many works have shown that approaches exploiting deep features achieve higher performance than radiomics \cite{lisson2022deep, sun2020deep, truhn2019radiomic, wei2021radiomics}. In our case, considering the evaluation on the external validation, CNN-derived features enabled higher accuracy compared with ViT-derived (\ref{tab:performance_test}). Regardless of datasets, ResNet maintains high accuracy values. 

\paragraph{In terms of explainability -} Saliency maps generated through GradCAM, EigenCAM, and ScoreCAM can effectively emphasize meaningful areas for prediction. While they serve as valuable visual and qualitative aids for prediction analysis, two primary concerns are presented: \textit{i)} each method yields distinct explanations, indicating a need for cautious interpretation; \textit{ii)} these methods can lead to inter-operator variability evaluation, as individual physicians may interpret the explanations differently; \textit{iii)} these explanations are inherently local, posing challenges to extract global insights. In contrast, radiomic features offer clearer and more comparable explanations. Since it is possible to associate each radiomic feature with meaning/behavior related to its analytical definition, they enable the comparison of features identified in other studies with those observed in clinical practice.

\paragraph{Comparing CNN-derived and ViT-derived features -} As highlighted in Section~\ref{ch:results_method}, a notable alignment was observed between CNN-derived and radiomic features, whereas none was found with those extracted from ViT. This outcome aligns intuitively with an examination of the internal structures of the two architectures.
CNN-based approaches typically rely on hierarchical feature extraction through a series of convolutional and pooling layers, capturing local patterns and spatial dependencies within the image. In contrast, ViT architectures adopt a self-attention mechanism, enabling them to capture global context by considering relationships between all image patches simultaneously. This leads CNNs to excel at capturing fine-grained details and local structures while ViTs in modeling long-range dependencies and contextual understanding. The authors explicitly admit that some intrinsic biases present in convolutional neural networks—such as translation equivariance and locality—are absent from transformers \cite{dosovitskiy2020image}. This similarity between radiomic features and CNN-derived features arises from their shared principle of extracting local characteristics. Convolutional filters in CNNs capture local information through the Receptive Field kernels and a hierarchical structure of layers. Likewise, radiomic features are predominantly derived from matrices that analyze the relationships among gray levels of adjacent pixels. This common approach aligns CNN-derived features closely with radiomic features, setting them apart from ViT-derived features. For these reasons, we consider \acrshort{MethodAcronym} as a global explanation method to explain CNN-derived features.

\subsection{Explainability Properties of \acrshort{MethodAcronym}}

\acrshort{MethodAcronym} offers a distinct perspective by explaining deep features through associations with the intrinsic interpretable radiomic features. This approach enables a richer clinically relevant understanding that surpasses the limitations of activation map-based methods.
The advantages of \acrshort{MethodAcronym} can be summarized as follows:

\begin{itemize}
    \item \textbf{provides global explanations:} it enables global insights, facilitating the extraction of clinically meaningful patterns across the entire dataset. This global perspective is valuable for clinical applications, where understanding overarching trends is as critical as individual predictions;
    \item \textbf{uses intelligible features:} the interpretable nature of radiomic features allows radiologists to correlate them with established clinical and radiological knowledge. While activation maps highlight regions of interest and enable only a qualitative analysis, radiomic features offer quantitative insights, helping bridge model predictions with well-known clinical evidence;
    \item \textbf{maintains model accuracy without training constraints:} approaches try to overcome the explainability-accuracy trade-off (e.g., Self-Explaining Neural Networks) often encounter a trade-off between robustness and accuracy due to imposed architectural constraints \cite{alvarez2018towards}. In fact, a change in test accuracy was shown as robustness regularization increases \cite{elbaghdadi2020self}. As a result, these model categories increase robustness and explainability at the cost of accuracy. Other methods require a specific structure for their use. For example, traditional activation-based techniques such as CAM \cite{zhou2016learning}, necessitate specific architectural configurations, like a Global Average Pooling layer before dense layers, limiting architecture flexibility. In contrast, \acrshort{MethodAcronym} allows the optimization of any CNN model using a standard training process. Then, in a \textit{post-hoc} manner, the correlation between deep and radiomic features is computed, thus preserving model performance and flexibility across tasks. For this reason, \acrshort{MethodAcronym} \textit{i)} achieves transparency without compromising predictive accuracy, \textit{ii)} mitigates some concerns related to the explainability-accuracy trade-off, and \textit{iii}) overcomes the accuracy-explainability trade-off dilemma.
\end{itemize}

\subsection{Mitigating XAI Concerns Through \acrshort{MethodAcronym}}

Explainable AI is a challenging branch that certainly provides substantial advantages over traditional AI-based approaches for X-CDSS development and implementation. However, although there has been a tendency to integrate global and local model explanations into the conventional development pipeline, XAI has also raised some criticisms. In fact, it seems that in some circumstances, an explanation may have also negative effects: \textit{i)} if the model displays less information, then excluding an explanation, users are not confused about the model (lower cognitive overload); \textit{ii)} often, intrinsic explainability is weak, and users should be trained to understand the explanations; \textit{iii)} it appears that end-users are more interested to understand \emph{'what the system does'} rather then \emph{'how the system works'} \cite{bell2022s}.
However, some of the mentioned issues can be mitigated considering \acrshort{MethodAcronym} and are discussed below.

\paragraph{An Explanation Makes the Model Reliable?} In our research, the significant inconsistencies in methods relying on saliency maps were confirmed,  a finding corroborated by numerous other studies \cite{cerekci2024quantitative, prinzi2024yolo, Zhang2022overlooked}. Consequently, even though these methods can identify where a model focuses for prediction, their reliability remains doubtful \cite{cerekci2024quantitative, oh2020deep, prinzi2023explainable, prinzi2024yolo, signoroni2021bs, Zhang2022overlooked}. As a result, it is impossible to state with certainty the reliability of the trained model and to extract certain clinical insights. Furthermore, the need to propose new methods beyond the visualization map-based methods was highlighted.
\cite{papanastasiou2023attention}.

In light of this, we conducted a qualitative comparison of \acrshort{MethodAcronym} with three widely used saliency map generation methods, to prove \acrshort{MethodAcronym} reliability (see Figure~\ref{fig:saliencyMaps}). The exploited techniques in the comparison are among the most representative and widespread in the literature for explaining deep architecture predictions: GradCAM, is representative of a gradient-based method, while ScoreCAM and EigenCAM avoid gradient computation to overcome some GradCAM shortcomings. In this perspective, our results highlight the unreliability of the saliency map methods and comply with previous papers \cite{Zhang2022overlooked, prinzi2024yolo}. In particular, regarding Figure~\ref{fig:saliencyMaps} we can highlight two main concerns:

\begin{itemize}
    \item \textit{discrepancy}: the three methods return different explanations. In particular, for the image in Row 1, all the explanations differ. For the other images (Rows 2, 3, and 4), GradCAM explanation is deeply different from ScoreCAM and EigenCAM. This can negatively affect physician interpretation;
    \item \textit{unreliability}: although the first two images (Rows 1 and 2) show correct predictions, explanations seem unreliable. In fact, saliency maps highlight the background as the most important part.
\end{itemize}

\acrshort{MethodAcronym} allows medical professionals to analyze whether radiomic features, whose significance is well defined, are aligned with the clinical literature. In addition, it is possible to extract new clinical findings, such as those discussed in Section~\ref{ch:clinical_validation}. This aspect overcomes the human tendency towards a positive interpretation of results by choosing the solution they want to be the correct one \cite{Bornstein2016}, providing an objective explanation. 

\paragraph{Accuracy and Explainability trade-off Dilemma -} Achieving the trade-off between explainability and accuracy is still an open dilemma \cite{bell2022s}. Although this trade-off has been widely discussed, there are some experiments where it is highlighted that in the clinical context, a patient would prefer an accurate model rather than an explainable one \cite{van2021trading}. This would suggest that one should never sacrifice accuracy for the sake of explainability.
Considering the absence of a definitive solution to this question, \acrshort{MethodAcronym} refrains from imposing constraints on model training. Instead, it provides an approach where radiomic features, although less predictive than deep ones, are exclusively utilized for explanatory purposes. It remains a task of the deep model to perform prediction accurately.

\paragraph{Explaining apparently incomprehensible patterns -} The abstraction mechanism within convolutional networks implies that the extracted features have a significantly higher level of abstraction than radiomic features. In this perspective, XAI methodologies must recognize that, although it is possible to explain model decisions, understanding the relationships identified by deep architectures may be inherently incomprehensible. However, \acrshort{MethodAcronym} elucidates that a correlation exists between CNN-derived and radiomic features, demonstrating a convergence between physician insights and AI-based interpretations, as highlighted in the following section.

\subsection{Radiomics-based explanation and its clinical validation}
\label{ch:clinical_validation}

\begin{figure}[t]
    \centering
        \includegraphics[width=0.5\textwidth]{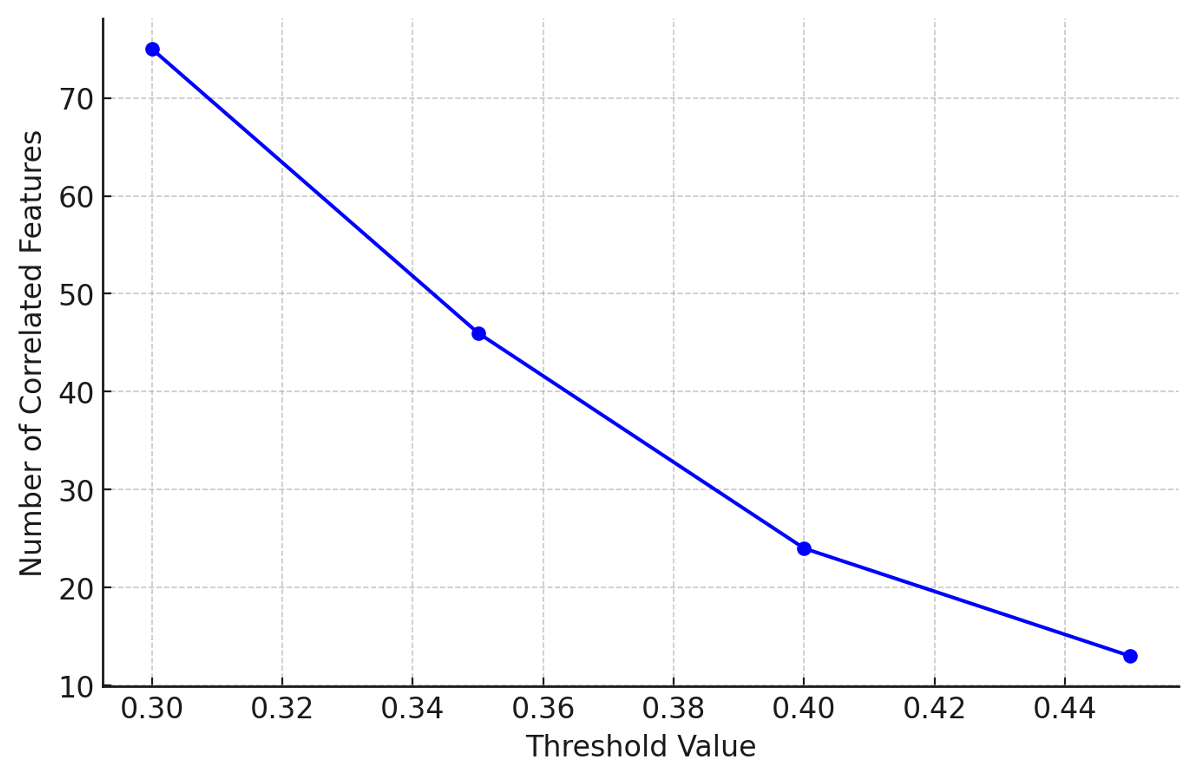}
        \caption{Correlation trend between radiomic and deep features.}
        \label{fig:correlationTrend}
\end{figure}

The radiomic features, found to be correlated with the CNN-derived features (Figure~\ref{fig:xai_corr}), have already been used in literature breast cancer studies, both diagnostic and prognostic. 
It is worth noting that if a study identifies a specific feature as important, it allows for direct comparisons to analyze if the same feature holds significance in other studies. This is not feasible with only qualitative saliency maps. Radiomic features, representing and quantifying the manifestation of the underlying tumor phenotype, can be considered biomarkers to be integrated into clinical practice. As a result, they offer a quantitative basis for influencing clinical decision-making.
For example, in the study by Cui \textit{et al.} \cite{cui2023radiogenomic} on the link between ultrasound radiomic features and biological functions in the prediction of HER2 status in breast cancer, Size Zone Non-Uniformity (SZNU) feature was associated with 1,871 genes in 75 perturbations and with carbon metabolism in cancer regarding biological functions; another radiomic feature analyzed in the same study \cite{cui2023radiogenomic} was Run Length Non-Uniformity (RLNU), associated with the cell cycle and intercellular communication. In the study by Youk \textit{et al.} on the analysis of the ultrasound radiomic features in benign and malignant breast masses \cite{youk2020grayscale}, 22 grayscale radiomic features were selected, some of which are also used in our study, such as Kurtosis, Energy, Run Length non-uniformity. 
The \textit{energy}-derived features such as \textit{Energy} and \textit{Total Energy} are closely related to the intensity of gray levels. These features were found crucial in several studies related to breast cancer, such as the proposed one, and in DCE-MRI, in which a strong correlation is present with the temporal evolution of the contrast agent \cite{prinzi2024breast}.
These radiomic features describe the change in gray intensity after administration of contrast medium, which is more rapid in malignant lesions, so these latter shows a more rapid increase in energy, \textit{viceversa} for benign lesions. In breast ultrasound evaluation, excluding completely anechoic lesions, gray levels related to the echogenicity of a focal lesion may vary, but most malignant lesions are hypoechoic \cite{rahbar1999benign}.
Although the correlation between gray level intensity and the malignant lesion is different in DCE-MRI than in ultrasonographic features, even in ultrasonographic radiomic analysis, we can assert that both Energy and Total Energy emerge as characterizing features in ultrasonography, thereby laying a solid foundation for subsequent clinical investigations.
The significance of signal intensity in ultrasound has been demonstrated in various anatomical districts \cite{papini2002risk, tessler2017acr}, indicating its potential importance for breast imaging as well.
Another example is the correlation between lesion heterogeneity, habitats' presence, and malignancy \cite{marusyk2010tumor,diaz2012tumor}. The entropy is a quantity that measures the degree of disorder in a system. Among radiomic features, \textit{entropy} is directly proportional to the heterogeneity of the lesion 'system' (\textit{e.g.} due to prominent vascularization and presence of different types of cells) and thus its malignancy.

As a consequence, \acrshort{MethodAcronym}, by establishing a link between deep features and related radiomic features that can be interpreted from a radiological point of view, can reduce the gap between black-box models and the explainability properties required by radiologists. Our results confirm an interesting overlap between the radiomics features and radiologist insights. This enables deep features to be considered clinical biomarkers such as radiomic features, with the advantage of obtaining more performing models. The missing link in this chain - the correlation between deep features and radiomics - is provided by \acrshort{MethodAcronym}. To provide more detail on this last aspect, Figure~\ref{fig:correlationTrend} depicts the correlation trend.

\subsection{Clinical Implications of \acrshort{MethodAcronym} Use}
\label{ch:implications}

\acrshort{MethodAcronym} contributes to improving current clinical practices in the radiologic domain. In particular, three main aspects were addressed:

\begin{itemize}
    \item \textbf{model clinical validation through quantitative radiomic features:} \acrshort{MethodAcronym} stands out for its focus on clinical validation, using quantitative radiomic variables that provide an objective and measurable view of image global characteristics, allowing for more accurate and reproducible radiologist-based assessment. This aspect is essential for model reliability;
    \item \textbf{reduction of inter-operator variability in interpreting model findings:} \acrshort{MethodAcronym} improves model explanation, overcoming the qualitative interpretation of saliency maps, which may be subject to inter-operator variability. As a result, \acrshort{MethodAcronym} enhances clinicians' trust and facilitates the adoption of predictive models into everyday workflows;
    \item \textbf{integration of intelligible features within the clinical practice:} The integration of radiomic features into the interpretation of medical images provides an added layer of confidence in results validation. Radiologists can now combine traditional image analysis with the richness of radiomic features. This fusion enhances clinical decision-making and enhances the traditional reporting process with new possible features to use as biomarkers.
\end{itemize}

\subsection{\acrshort{MethodAcronym} Limitations}
\label{ch:limitations}

From a methodological perspective, \acrshort{MethodAcronym} application requires ROIs segmentation for radiomic feature extraction. In contrast, deep learning models typically automatically learn the most important features (the ROI is not needed). In particular, for \acrshort{MethodAcronym}, it is crucial to have pre-defined segmentation of ROIs to ensure that radiomic features are extracted solely from pathological areas, avoiding irrelevant information from surrounding healthy tissue. However, although segmentation is a resource-intensive process, several tools are available nowadays for accurate and rapid segmentation.
From a clinical perspective, it is essential for radiologists to understand the meaning of each radiomic feature to derive clinical insights accurately. While the number of extracted features is often substantial, radiomic features have become increasingly standardized and are now being widely incorporated into clinical literature. Furthermore, feature extraction software is typically accompanied by detailed documentation, making it easier for non-experts to comprehend and utilize these features effectively.

\section{Conclusion}
\label{sec:Conclusion}

In this study, we have presented \acrshort{MethodAcronym}, a novel approach for the development of eXplainable Clinical Decision Support Systems, bridging the gap between accuracy and explainability without imposing methodological constraints on model performance. Through a comprehensive analysis comparing radiomic, Convolutional Neural Network (CNN)-derived, and Vision Transformer (ViT)-derived features in the context of breast cancer diagnosis, we have shed light on several critical aspects of XAI and its application in clinical decision-making.

Our findings highlight significant differences between radiomic, CNN-derived, and ViT-derived features in terms of accuracy and explainability. While deep architectures, particularly CNNs, demonstrate superior performance in terms of accuracy, radiomic features offer clearer and more comparable explanations due to their well-defined significance. \acrshort{MethodAcronym} revealed a notable alignment between CNN-derived and radiomic features, indicating shared principles of extracting local characteristics. Moreover, we addressed concerns surrounding XAI through \acrshort{MethodAcronym}, offering insights into its reliability, the accuracy-explainability trade-off dilemma, and the complexity of deep model explanations. By identifying radiomic features that exhibit a high correlation with CNN-derived features we were able to draw some important clinical conclusions. The correlation between these features and those observed in previous studies on breast cancer diagnosis underscores the clinical relevance and potential utility of our findings.

A promising future research direction would be exploring how this approach could be extended to explain features derived from ViTs, given the different feature extraction processes compared to CNNs. Additionally, incorporating structured semantic information, such as BI-RADS attributes (\textit{e.g.}, density, margin shape, microcalcifications, etc.), could further enrich model interpretability. These attributes could serve dual purposes: \textit{i)} improving classification accuracy and \textit{ii)} enabling the association of global intelligible features with extracted deep features. This exploration could open up new pathways for aligning deep learning features with clinically recognized descriptors, facilitating more meaningful insights in medical image analysis.

Summarizing, the use of Radiomics in CNN explanation represents a promising advancement in the development of XCDSSs. It provides a new perspective on XAI methods that deviate from the methods based on visualization maps. In addition, it provides a form of global explanation, an aspect often overlooked in current research on explainable methods in medical imaging.

\subsection*{Ethics Statement}
\label{sec:EthicsStatement}

The study was approved by the Ethical Committee of the Policlinico University Hospital ‘P. Giaccone’ of Palermo (minute N. 02/2019 of 18/02/2019). Informed consent was waived because of the retrospective nature of the study and the analysis. The data of the patients involved were processed according to the institutional privacy regulations of the Policlinico University Hospital ‘P. Giaccone’ of Palermo. The privacy rights of the human subjects involved in the study are respected because anonymized clinical and imaging data were used.

\section*{CRediT authorship contribution statement}

\textbf{Francesco Prinzi}: conceptualization, methodology, software, formal analysis, writing original draft, visualization; \textbf{Carmelo Militello}: conceptualization, methodolog\textbf{}y, formal analysis, writing original draft, visualization; \textbf{Calogero Zarcaro}: validation, investigation, data curation; \textbf{Tommaso Vincenzo Bartolotta}: validation, investigation, data curation, supervision; \textbf{Salvatore Gaglio}: validation, investigation, writing - review and editing, supervision; \textbf{Salvatore Vitabile}: validation, investigation, resources, writing - review and editing, supervision, project administration, funding acquisition.

\section*{Funding}

This research was \textit{i)} funded by the Italian Complementary National Plan PNC-I.1 "Research initiatives for innovative technologies and pathways in the health and welfare sector” D.D. 931 of 06/06/2022, "DARE - DigitAl lifelong pRevEntion" initiative, code PNC0000002, CUP: B53C22006460001; and \textit{ii)} funded by the European Union – Next Generation EU - Progetti di Ricerca di Rilevante Interesse Nazionale (PRIN) 2022, Prot. 2022ENK9LS. Project: "EXEGETE: Explainable Generative Deep Learning Methods for Medical Image and Signal Processing", CUP: B53D23013040006.

\section*{Acknowledgments}
\label{sec:Acknowledgement}

The authors would like to thank Dr. Alice Latino for their substantial contribution to some experiments.


\bibliographystyle{unsrtnat}
\bibliography{references}  

\end{document}